\documentclass[10pt]{article}

\usepackage[utf8]{inputenc} 
\usepackage[T1]{fontenc}    
\usepackage{hyperref}       
\usepackage{url}            
\usepackage{booktabs}       
\usepackage{amsfonts}       
\usepackage{nicefrac}       
\usepackage{microtype}      

\usepackage{bm}
\usepackage{amsmath}
\usepackage{amssymb}
\usepackage{listings}

\usepackage[numbers]{natbib}
\usepackage{float}
\usepackage{graphicx}
\usepackage{rotating}
\graphicspath{{./}{../png/}{../eps/}{../asy/}}
\newcommand{\defwidth}{\textwidth}
\newcommand{\graphic}[2][width=\defwidth]{\centerline{\includegraphics[#1]{#2}}}
\newcommand{\optlabel}[1]{\ifx&#1& \else \label{#1} \fi}
\newcommand{\legend}[2][]{\makebox[\defwidth][c]{\begin{minipage}[t]{1.\defwidth}\caption{\normalsize{#2}}\optlabel{#1}\end{minipage}}\\}
\newenvironment{portrait}[1][!htp]{\begin{figure}[#1] \begin{centering}}{\end{centering}\end{figure}}

\title{Quantal synaptic dilution enhances sparse encoding and dropout regularisation in deep networks}

\author{%
  Gardave S. Bhumbra. \\
  Department of Neuroscience, Physiology and Pharmcology,\\
  UCL,\\
  Gower St., 
	London, WC1E 6BT,
	UK.\\
  \texttt{g.bhumbra@ucl.ac.uk} \\
}

\begin{document}

\maketitle

\begin{abstract}
	Dropout is a technique that silences the activity of units stochastically while training deep networks to reduce
	overfitting. Here we introduce Quantal Synaptic Dilution (QSD), a biologically plausible model of dropout
	regularisation based on the quantal properties of neuronal synapses, that incorporates heterogeneities in response
	magnitudes and release probabilities for vesicular quanta. QSD outperforms standard dropout in ReLU multilayer
	perceptrons, with enhanced sparse encoding at test time when dropout masks are replaced with identity functions,
	without shifts in trainable weight or bias distributions. For convolutional networks, the method also improves
	generalisation in computer vision tasks with and without inclusion of additional forms of regularisation. QSD also
	outperforms standard dropout in recurrent networks for language modelling and sentiment analysis. An advantage of QSD
	over many variations of dropout is that it can be implemented generally in all conventional deep networks where
	standard dropout is applicable. 

\end{abstract}

\section{Introduction}

Dropout is a regularisation method used in deep learning to reduce overfitting. The technique discretely masks the
outputs of hidden units stochastically setting a predefined proportion to zero. This avoids complex co-adaptation of
trainable weights within a hidden layer \citep{hinton2012improving,srivastava2014dropout}. Dropout effectively provides
a computationally efficient method to train a large number of different networks simultaneously since the number of
possible combinations of unmasked units grows exponentially with network size \citep{srivastava2014dropout}. This
results in a regularised average of many models at test time \citep{warde2014empirical,srivastava2014dropout}. 

From a neuroscience perspective, discrete masking of units is analogous to the failure of a nerve impulse to evoke the
release of a quantum of neurotransmitter at biological synapses. When an action potential reaches a pre-synaptic
terminal, the voltage-activated influx of calcium can fail to result in release of neurotransmitter contained within a
synaptic vesicle. Synaptic transmission is therefore also stochastic, and the probability of release $p$ is one of the
three fundamental input-independent quantal properties of a synapse. The other two are the number of contacts $n$, and
the quantal size $q$ that quantifies the magnitude of each post-synaptic response to a single vesicular quantum.

A binomial model \citep{kuno1971quantum} scaled by the quantal size $q$ can be used to express the mean quantal response
as the product the three quantal parameters: $npq$. Experimental recordings from central neurons show however that
synapses exhibit heterogeneities in the probabilities of release $p$ \citep{walmsley1988nonuniform} and quantal sizes
$q$ \citep{lisman1993quantal}. While such heterogeneities do not strongly impact upon the mean quantal response, they
contribute to the variance; a greater diversity in quantal size increases quantal response variances, whereas increased
heterogeneity in the probabilities of release results in a decrease \citep{silver2003estimation}.

Here we introduce quantal synaptic dilution (QSD), a biologically plausible model of dropout that incorporates both
sources of synaptic heterogeneity. QSD is implemented in a range of different deep architectures to compare with
standard dropout. In contrast to numerous existing variations of standard dropout, QSD is as generally applicable to
conventional deep networks and we show it improves regularisation in multilayer perceptrons, convolutional networks, and
recurrent networks.

\section{Quantal Synaptic Dilution}

`Dilution' refers to partial connectivity between two network components, and the term is used in neuroscience to
quantify the synaptic connectivity between two brain regions \cite{rolls2012advantages}. `Strong dilution` means that
the partial connectivity is permanent, whereas in `weak dilution' a fixed proportion of connections are transiently
removed at random \citep{hertz1991introduction}. Dropout is a form of weak dilution that rescales activations
\citep{labach2019survey} in order to approximate the average outputs of layers during training to the mean activations
at test time when the dropout masks are replaced with identity functions. 

While there are various versions of dropout that are mathematically equivalent, here we describe `inverted dropout'
since it is implemented in many deep learning libraries. The activation $\mathbf{a}$ of a layer of $n$ units with a
transfer function $g$, trainable weights $\mathbf{W}$ and biases $\mathbf{b}$, receiving input $\mathbf{x}$ is given by:

\begin{align}
	\mathbf{a} &= g(\mathbf{Wx}+ \mathbf{b}) \label{eq:test}
\end{align}

For an inverted dropout implementation at a given dropout rate $d$, the output $\mathbf{y}$ is the Hadarmard product of
the activations $\mathbf{a}$ with a data-independent vector $\mathbf{c}$ of coefficients $(c_1, c_2, \ldots, c_n)^T$.
The coefficients of $\mathbf{c}$ are evaluated from the product of a Boolean mask vector $\mathbf{m}$, where $m_i$ is
the $i$th Bernoulli sample with the retain probability $p=1-d$, and the rescaling size $q=1/p$.

\begin{align}
	\mathbf{y} &= \mathbf{c} \odot g(\mathbf{Wx} + \mathbf{b}),
	&\mathbf{c} = q\mathbf{m},
	\phantom{M}q = \frac{1}{p},
	\phantom{M}&m_i \sim \textrm{Bernoulli}(m \vert p)
	\label{eq:train}
\end{align}

A rescaling size of $q=1/p$ ensures an expected mean for the coefficients $\mathbf{c}$ of 1, and thus warrants
replacement of the dropout transformation with an identity function (or Equation \ref{eq:test}) at test time. Standard
dropout assumes a single value for the dropout rate $d$ and therefore also for the retain probability $p$. Heterogeneous
release probabilities however have been observed at central synapses \citep{walmsley1988nonuniform}, with their
distribution at hippocampal synapses fitted initially with a gamma probability density function
\citep{murthy1997heterogeneous}. Since probabilities are bounded in the range $[0,1]$, more recent methods of quantal
analysis \citep{silver2003estimation,bhumbra2013reliable} model heterogeneous release probabilities using the closely
related beta distribution:

\begin{align}
\mathcal{P}(p \vert \alpha, \beta) &= \frac{p^{\alpha-1}(1-p)^{\beta-1}}{B(\alpha, \beta)},
	\phantom{i}\alpha, \beta > 0,
	\phantom{i}B(\alpha, \beta) = \frac{\Gamma(\alpha)\Gamma(\beta)}{\Gamma(\alpha+\beta)}, 
	\phantom{i}\Gamma(\gamma) = \int_{0}^{\infty} g^{\gamma-1}e^{-g}dg
\end{align}

where $\alpha$ and $\beta$ are the distribution parameters. Examples of beta distributions are shown in Figure
\ref{fig:beta}.

\begin{portrait}[h] 
	
\graphic[width=0.35\defwidth,trim={0cm 1cm 0cm 0.5cm}]{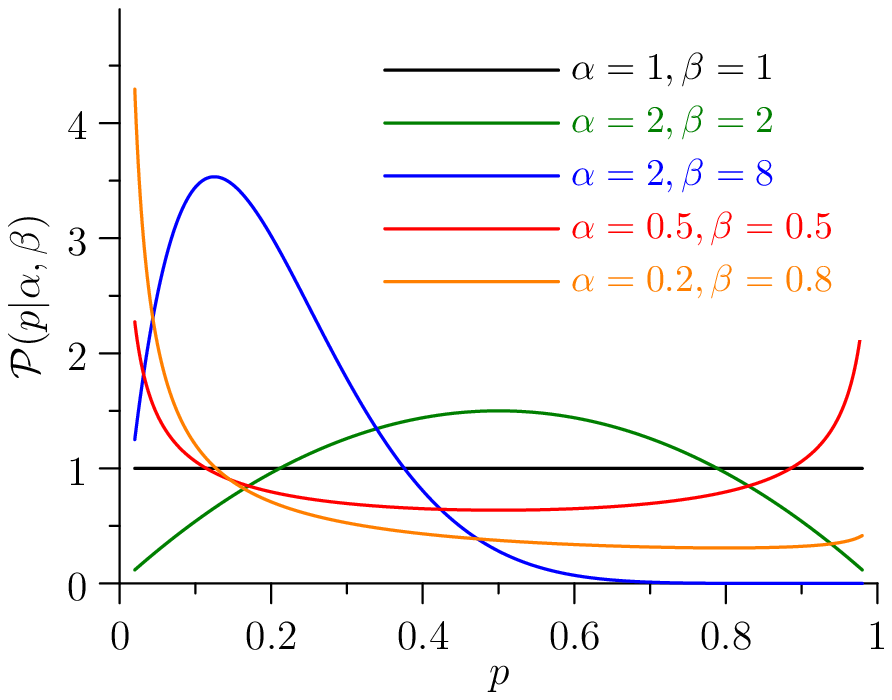} 

	\legend[fig:beta]{Beta distributions for different $\alpha$ and $\beta$; $\alpha=\beta=1$ corresponds to a uniform
	distribution.} 

\end{portrait}

The first heterogeneity of QSD is modelled by distributing the retain probabilities $p_i$ used for the Bernoulli
sampling according to a beta distribution:

\begin{align}
	m_i \sim \textrm{Bernoulli}(m \vert p_i),\phantom{M} &p_i \sim \mathcal{P}(p \vert \alpha, \beta)
\end{align}

The distribution parameter $\alpha$ represents a homogeneity hyperparameter. Given a mean retain probability $\bar{p}$,
a corresponding value for $\beta$ is determined by the functional relation: $\beta = \alpha (1-\bar{p})/\bar{p}$.

The second heterogeniety of QSD is modelled by distributing the rescaling size $q$, that is analogous to the quantal
size. As for inverted dropout, the expected mean for the rescaling sizes corresponds to the reciprocal of the mean
retain probability $1/\bar{p}$. Since the retain probabilities $\mathbf{p}$ are already beta-distributed with a mean
$\bar{p}$, it is efficient to compute the rescaling element $q_i$ for the $i$th unit by dividing its corresponding
retain probability $p_i$ by the square of the mean retain probability $q_i = p_i/(\bar{p})^2$, giving an expected mean
for $\mathbf{q}$ of $1/\bar{p}$. 

Combination of the beta-distributed Bernoulli masks with the rescaling sizes results in expectation scaling lower bound
of layer outputs of 1 (see Appendix). The linear relation between the retain probabilities $\mathbf{p}$ and rescaling
sizes $\mathbf{q}$ is consistent with biological observations since strong positive correlations between release
probabilities and quantal sizes have been demonstrated at synapses of pyramidal cells in the visual cortex
\citep{hardingham2010quantal} with correlation coefficients as high as $r=0.89$ in layer V. Combining the two
heterogeneities together, the QSD transformation can be expressed in a form similar to Equation \ref{eq:train}:

\begin{align}
	\mathbf{y} &= \mathbf{c} \odot g(\mathbf{Wx} + \mathbf{b}),
	\phantom{i}c_i = q_i m_i,
	\phantom{i}q_i = \frac{p_i}{(\bar{p})^2},
	\phantom{i}m_i \sim \textrm{Bernoulli}(m \vert p_i),
	\phantom{i}p_i \sim \mathcal{P}\left(p \vert \alpha, \alpha \left( \frac{1-\bar{p}}{\bar{p}} \right)\right)
\end{align}

As for inverted dropout, the QSD employs an identity function at test time. An increase in the homogeneity parameter
$\alpha$ decreases the variance of $\mathbf{p}$ and consequently $\mathbf{q}$, tending to a standard dropout model as
$\alpha \to \infty$. When $\alpha$ is finite, the outputs of units distributed with high retain probabilities are
amplified relative to those at low probabilities. Resulting effects on the distributions of outputs are investigated in
the present study (Figure \ref{fig:mnistmlppq}). Programming implementation and more comprehensive consideration of QSD
scaling effects are detailed in the Appendix.

\subsection{Related work}

Dropout has attracted considerable interest particularly in the fields of variational Bayesian analysis and computer
vision with many variants proposed. DropConnect \citep{wan2013regularization} randomly masks network weights rather than
activations. Fast dropout is a Bayesian variant of dropout \citep{wang2013fast} that uses approximate Gaussian sampling
obviating boolean masks and thus avoids limiting the proportion of active and therefore trainable units. Variational
dropout \cite{kingma2015variational} models Gaussian multiplicative noise to derive a stochastic gradient variational
Bayesian (SVGB) inference algorithm. 

From a Bayesian perspective, dropout may be regarded as a form of variational inference for deep Gaussian processes to
model uncertainty minimising Kullback-Leibler \citep{gal2016dropout} or alpha- \cite{li2017dropout} divergences.
Concrete dropout \citep{gal2017concrete} is a variant that replaces the discrete dropout masks with a continuously
sampled concrete discribution \citep{maddison2016concrete}. \citet{partaourides2017deep} propose a hierarchical Bayesian
model of Black-Box Variational Inference (BBVI) based on DropConnect imposing a beta prior over network weights. In soft
dropout \citep{xie2019soft}, a beta prior is approximated using a half-Gaussian or half-Laplacian to derive a SVGB
algorithm with adaptive dropout rates.

Many dropout variants specific to convolutional networks have been proposed. In `Maxout' networks
\citep{goodfellow2013maxout}, the dropout masks are activity-dependent, whereby maxout feature maps are selected from
the maximum outputs across a defined number of feature maps. `Spatial dropout' \citep{tompson2015efficient} increases
the granularity of dropped components from individual units to feature maps. The use of batch normalisation
\citep{ioffe2015batch} in convolutional networks \citep{he2016deep} has impacted upon the use of dropout. While the
rescaling used in dropout approximates the average outputs to the mean activations during test time, a side effect of
`variance shift' \citep{li2019understanding} on batch normalisation may adversily affect predictions at test time,
although mitigation strategies have been proposed \citep{cai2019effective}.

This novelty of the present study is that it is the first to investigate a biologically realistic model of dropout at
the quantal synaptic level. QSD is not a Bayesian variant of dropout based on variational inference. It is therefore
compared with standard dropout for three conventional deep network architecture types: multilayer perceptrons,
convolution networks, and recurrent networks. Performances of dropout experiments are inherently variable, and therefore
the focus of this study has been on reporting reproducible experiments rather than comparing best performances. For all
experiments, eight training runs were performed with random seeds matched across all different conditions.  Performance
metrics for each run were computed by averaging the evaluations from the final three epochs. Comparisons of performances
across different conditions were confirmed statistically using unpaired Wilcoxon rank sum tests.

\section{Multilayer perceptron dilution}

Fully connected multilayer perceptron networks with three hidden layers of 128, 256, and 512 ReLU units were trained on
the MNIST data set (60000 training and 10000 test samples of 28x28 images with 10 classes) with a softmax 10-unit output
layer. Cross-entropy costs were minimised using a stochastic gradient descent optimiser with momentum $(\mu=0.9)$ and a
batch size of 64 for 100 epochs. The initial learning rate of 0.01 was multiplied by 0.2 on epochs 30, 60, then 80.
Weights were initialised using variance scaling \citep{he2015delving}. Dropout or QSD masks were positioned after each
of the three ReLU non-linearities at identical rates. No other regularisation or data augmentation was used.

Learning curves (Figure \ref{fig:mnistmlp}, top) show a reduced descent in training cost with a dropout rate of 0.2, but
little difference between standard dropout and QSD with the homogeneity parameter $\alpha$ assessed over the range
$[0.2, 5]$. The curves for test costs however differ, with the final median performance for QSD at $\alpha=0.2$ of 0.061
nats is significantly less than for standard dropout (0.072 nats, $Z=3.36$, $P<0.001$). Increases in $\alpha$ raises the
test cost towards standard dropout performances. Group results (Figure \ref{fig:mnistmlp}, bottom) show that standard
dropout over the range $[0, 0.5]$ does not approach the test performance of QSD ($\alpha=0.2$) at a rate of 0.2. When
$\alpha$ is low, large dropout rates (>0.2) elevates test costs as a result of impaired training (with training costs
>0.1 nats for $\alpha=0.2$ at a rate of 0.3); increases in $\alpha$ however alleviates this effect.

\begin{portrait}[!htp] 
	
	\graphic[width=0.9\defwidth,trim={0 0.1cm 0 0},clip]{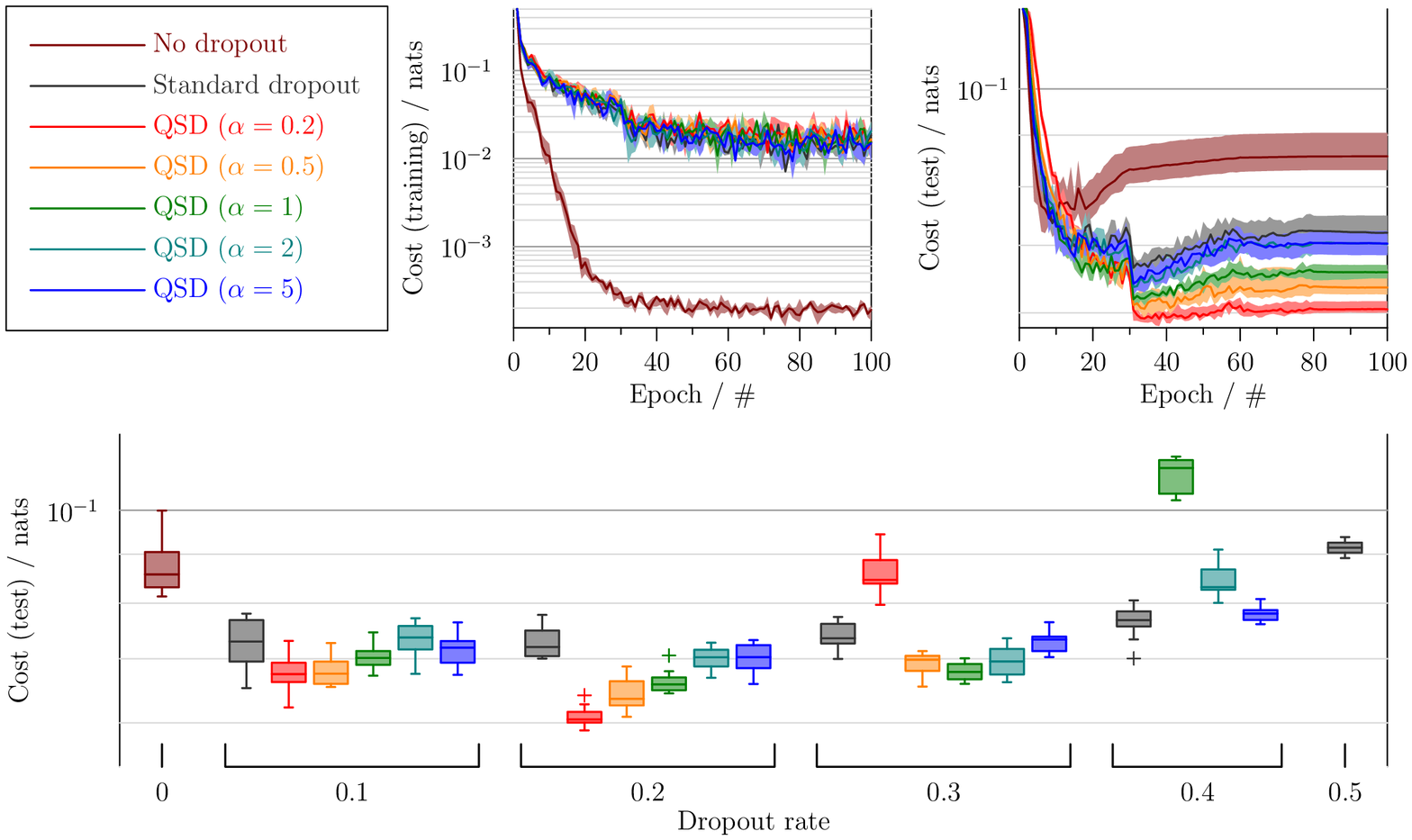} 

	\legend[fig:mnistmlp]{Multilayer perceptron with 3 layers, training on the MNIST data set with dropout at different
	rates and quantal synaptic dilution (QSD) for different values of $\alpha$. \textit{Top:} Learning curves show
	medians (lines) and inter-quartile ranges (bands) at a dropout rate 0.2 with standard dropout and QSD with different
	$\alpha$ values.  \textit{Bottom}: test performances after training at different dropout rates.}

\end{portrait}

Improved regularisation could result from distributed retain probabilities, distributed rescaling sizes, or the
combination as employed for QSD. In order to isolate the contributions, we selectively ablated one heterogeneity while
keeping ther other. Figure \ref{fig:mnistmlppq} illustrates the results of experiments in which we fixed either the
rescaling size $q$ to $1/\bar{p}$ or the retain probability $p$ to $\bar{p}$, while maintaining the distributed
heterogeneity in the other parameter. For $\alpha$ values of 0.2 and 0.5, the learning curves (Figure
\ref{fig:mnistmlppq}A) demonstrate superior performance of QSD over the two isolated conditions. At $\alpha=0.2$, the
median test cost for distributed retain probabilities (0.070 nats) and distributed rescaling sizes (0.073 nats) were
significantly greater than that for the combination (0.061 nats, $Z=3.36$, $P<0.001$). Similar results were obtained for
$\alpha=0.5$ (0.073 nats and 0.072 nats compared to 0.064 nats, $Z\ge2.94$, $P\le0.003$).

A beta prior imposed on network weights used for a hierarchical Bayesian BBVI variant of DropConnect
\citep{partaourides2017deep} may improve regularisation by inducing sparsity. In order to investigate effects of QSD on
sparse encoding, the outputs of hidden layers following training were evaluated for a forward pass of the test input
data. Figures \ref{fig:mnistmlppq}B-D show histograms of the outputs averaging over all samples for individual units
(left) and averaging over all units for individual samples (right). The results obtained from all experimental
are summarisedk in the Appendix. For Figure \ref{fig:mnistmlppq}B, the dropout masks were kept active as during
training. While the mean output across units shows increased sparsity for the distributed retain probability and QSD
conditions, there are no systematic differences among the different conditions when averaging over all units across
individual samples.

For a good sparse model there should only be a few number of highly activated units \citep{glorot2011deep}, with the
output of hidden layers exhibiting a low mean and variance of activations across different data samples
\citep{srivastava2014dropout}. When evaluating the outputs shown in Figure \ref{fig:mnistmlppq}C, masks were inactivated
as during test time. The output across units, particularly in the third layer, show a increase in sparsity for the QSD
layers compared to the other conditions. Averages over all units across individual samples however show that only the
QSD layers exhibit reductions in means and variances compared to standard dropout. The relative progression of the
reduction in mean activation across layers suggests that QSD facilitates propagation of sparse encoding along the
network.

Figure \ref{fig:mnistmlppq}D illustrates similar output histograms with the masks inactive but with the input pixels of
the test data randomly permutated to remove any meaningful signal while preserving the original input distribution. Only
the QSD layers systematically exhibit increased sparsity across units and relative reductions in mean and variance of
outputs across samples compared to standard dropout. Once again the reduction in outputs for QSD along the network
compared to the other conditions suggests a facilitated propagation of sparse encoding, with a minimal activation of the
final hidden layer that is most consistent with an encoding of random inputs. These effects are not a result of changes
in the distributions of trainable variables, as evidenced by the lack of shifts in the histograms for the layer weights
and biases at the end of training (Figure \ref{fig:mnistmlppq}E). The improvement by QSD in sparse encoding and network
regularisation must therefore result from a dilution during training that combines heterogeneous retain probabilities
with distributed rescaling sizes.

\begin{portrait}[!htp] 
	
	\graphic[width=1.5\defwidth,trim={0 0.1cm 0 0},clip]{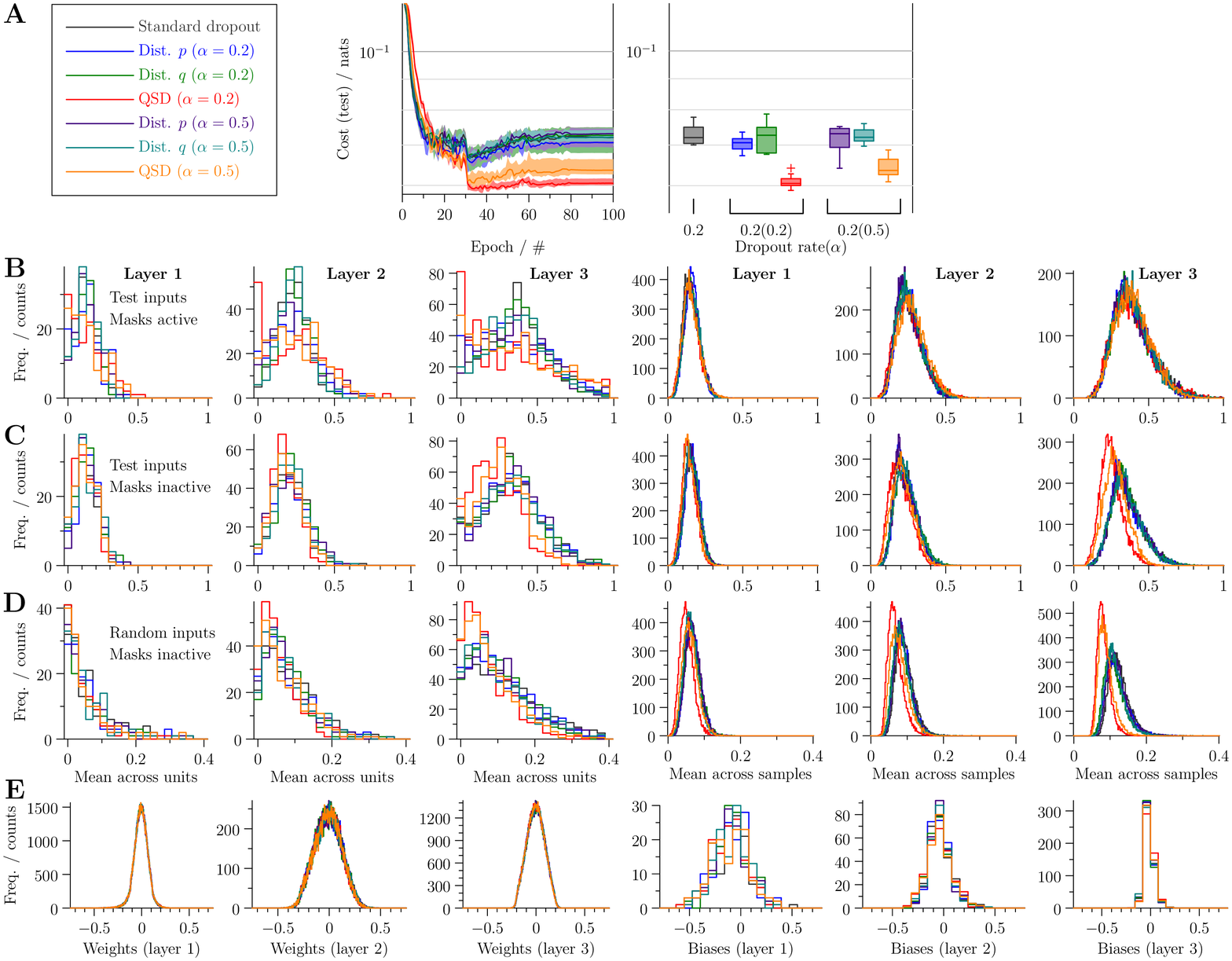} 

	\legend[fig:mnistmlppq]{Comparisons of standard dropout, distributed retain probabilities (Dist. $p$), distributed
	quantal sizes (Dist. $q$), and QSD. \textbf{A:} Learning curves and test performances for the different conditions.
	\textbf{B}: Distributions of layer outputs after training, forward passing test inputs with the dropout masks active
	averaging over all samples for individual units (\textit{left}) and over all units for individual samples
	(\textit{right}). \textbf{C}: The same outputs for a forward pass of the test data at test time with masks inactive. 
	\textbf{D}: The same outputs forward passing the randomised data with masks inactive (note reduced $x$-axis scale).
	\textbf{E}: Distributions of weights (\textit{left}) and biases (\textit{right}) for corresponding layers after
	training.}

\end{portrait}

\section{Convolutional network dilution}

A LeNet-5 convolutional architecture \citep{lecun1998gradient} was used to compare QSD with standard dropout on the
MNIST dataset. The network, comprising two ReLU convolutional layers alternating with two average pooling layers, two
hidden fully-connected ReLU layers, and an output softmax layer, was trained using a Nesterov-accelerated momentum
($\mu=0.9$) stochastic gradient descent optimiser to minimise the cross entropy. The batch size, training epochs, and
learning rate schedule were identical to those described in the previous section. Dropout or QSD masks were positioned
after each of the two ReLU non-linearities following the two convolutional layers and the two fully-connected hidden
layers with identical dropout rates. No other regularisation or data augmentation was used.

Learning curves (Figure \ref{fig:cnn}, top) illustrate that dropout at a rate of 0.1 improves generalisation.
Comparisons of standard dropout with QSD ($\alpha=0.2$) show improved performance of QSD (median test cost = 0.014 nats,
median test error = 0.0050) over standard dropout (test cost = 0.016 nats, error = 0.0055) with statistical significance
for test cost ($Z=3.26$, $P=0.001$) and error ($Z=2.42$, $P=0.016$). Group results (Figure \ref{fig:cnn}, middle) show
that standard dropout over the range $[0, 0.3]$ does not approach the test performance of QSD ($\alpha=0.2$) at a rate
of 0.1. Large dropout rates (>0.15) results in an elevated test cost when $\alpha$ is low, but again this effect is
alleviated by increases in $\alpha$.

\begin{portrait}[!htp] 
	
	\graphic[width=0.8\defwidth,trim={0 4cm 0 0},clip]{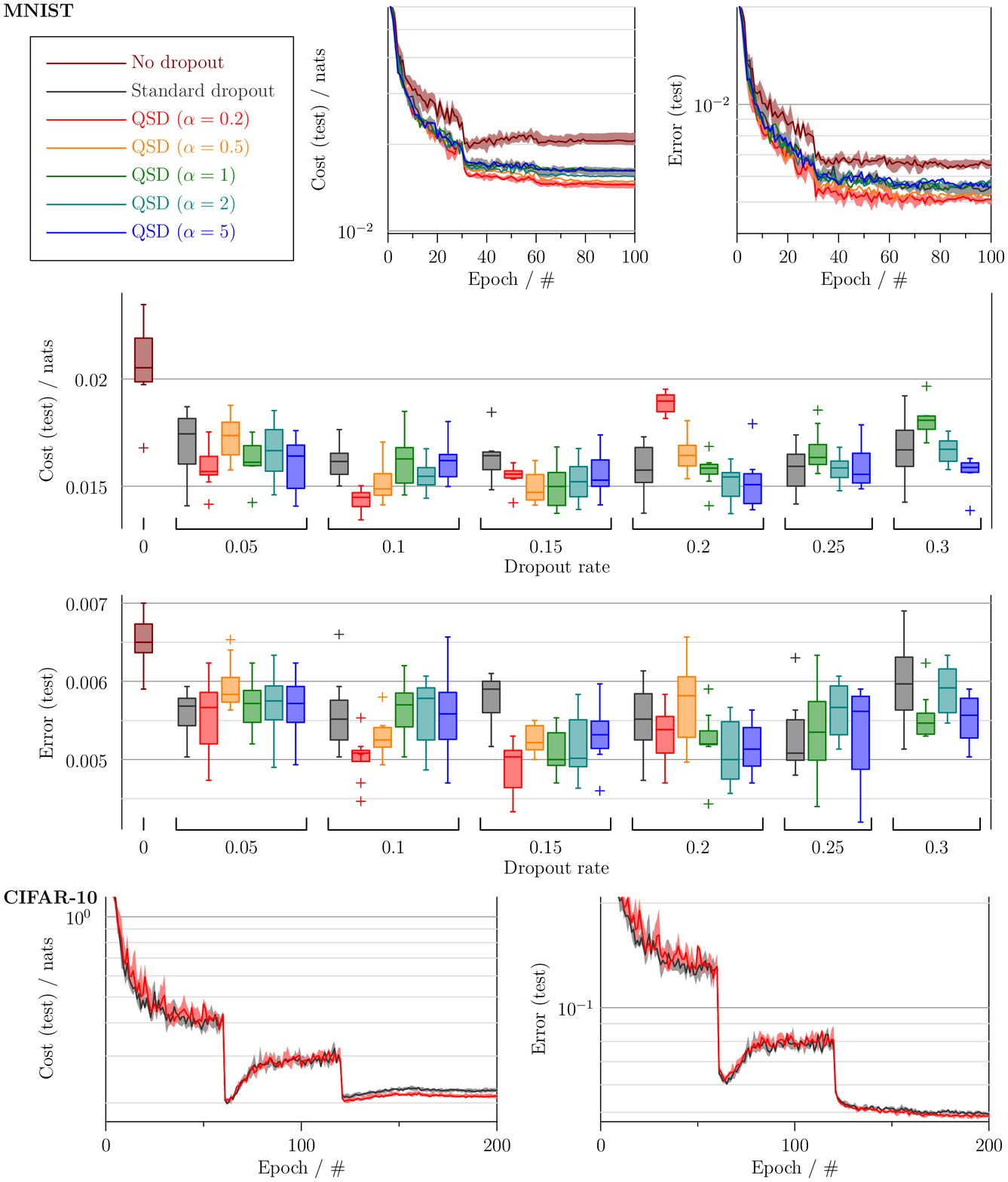} 

	\legend[fig:cnn]{Comparisons of dropout performances after training convolutional networks. \textit{Top}: Learning
	curves for LeNet-5 network architecture training on MNIST with a dropout of 0.1 (for QSD, the inter-quartile range is
	shown only for $\alpha=0.2$). \textit{Middle}: Test performances after training at different dropout rates.
	\textit{Bottom} Learning curves for 16-layer Wide ResNet ($k=6, N=2$) architecture for CIFAR-10 images with a dropout
	rate of 0.1.}

\end{portrait}

In order to compare QSD with standard dropout for more complex convolutional architectures with additional forms of
regularisation, a 16-layer Wide ResNet ($k=6, N=2$) architecture \citep{zagoruyko2016wide} was trained on the CIFAR-10
data dataset (50000 training and 10000 test samples of 32x32x3 images, with 10 classes). Dropout blocks were implemented
as described previously \citep{zagoruyko2016wide}, with a rate of 0.1 selected on the basis of the `drop-neuron'
performances reported by \citet{cai2019effective}. Dropout masks were inserted between the ReLU non-linearity and
convolution, as proposed by \citet{cai2019effective} to mitigate the effects of variance shift on batch normalisation.
Based on the LeNet-5 performance on MNIST (Figure \ref{fig:cnn}), the QSD homogeneity $\alpha$ was set to 0.2 with no
hyperparameter search.

Training parameters were set to those previously described \citep{zagoruyko2016wide}: batch size 128, momentum 0.9, with
200 epochs and learning rate reductions from 0.1 to 0.02, 0.004, and 0.0008 at epochs 60, 120, and 160 respectively,
weight decay 0.0005, global contrast normalisation of images, and training data augmentation with random horizontal
flips and random crops following 2-pixel reflections across all 4 edges. Training runs comparing standard dropout and
QSD (Figure \ref{fig:cnn}, bottom) show despite increased complexity and additional regularisation of the network, QSD
outperforms standard dropout according to both test cost (standard dropout 0.222 nats, QSD 0.211 nats, $Z=2.21$,
$P=0.027$) and error (standard dropout 0.0495, QSD 0.0487, $Z=2.048, P=0.040$).

\section{Recurrent network dilution}

In order to investigate the potential utility of QSD beyond training ReLU units for computer vision tasks, it was
compared with standard dropout in recurrent networks for natural language processing applications. We base the first
comparison on the experiment of \citet{zaremba2014recurrent} on the Penn Treebank dataset, comprising of 887,521 words
in total, training a two-layer LSTM recurrent network applying dropout to non-recurrent connections. `Medium' and
`large' models are detailed in the study of \citet{zaremba2014recurrent}, comprising 650 and 1500 units with respective
dropout rates of 0.5 and 0.65 at non-recurrent connections. The medium model was unchanged for the present study but
at an initial learning rate of 1, the large model sometimes diverged in the first epoch. In order to maintain
reproducibility across runs and match random seeds across conditions, the large model was adjusted to 1300 units and a
dropout rate of 0.6 which avoided divergence using standard dropout.

At a dropout of 0.5 and a QSD homogeneity $\alpha$ of 1 (and therefore $\beta=1$), the corresponding quantal
heterogeneity function would have the distinct and appealing property of conforming to a uniform distribution.
Unfortunately this resulted in a small risk of network divergence in the first epoch. The QSD homogeniety $\alpha$
parameter was therefore set to the lowest value that resulted in no divergence (5 for medium and 10 for large), with no
further hyperparameter search.  For both medium (Figure \ref{fig:ptblstm}, top) and large (Figure \ref{fig:ptblstm},
bottom) models, the training perplexity for the QSD regularised network remains greater than that for the standard
dropout network, while in both cases the validation perplexity is less for the QSD network towards the end of training.
For the medium model, the median test perplexity for QSD of 81.8 was less than that for standard dropout (83.5,
$Z=3.36$, $P<0.001$). Similar results were observed for the large model (standard dropout 79.7, QSD 78.9, $Z=3.36$,
$P<0.001$).

\begin{portrait}[!htp] 
	
	\graphic[width=1.0\defwidth,trim={0 0 0 0},clip]{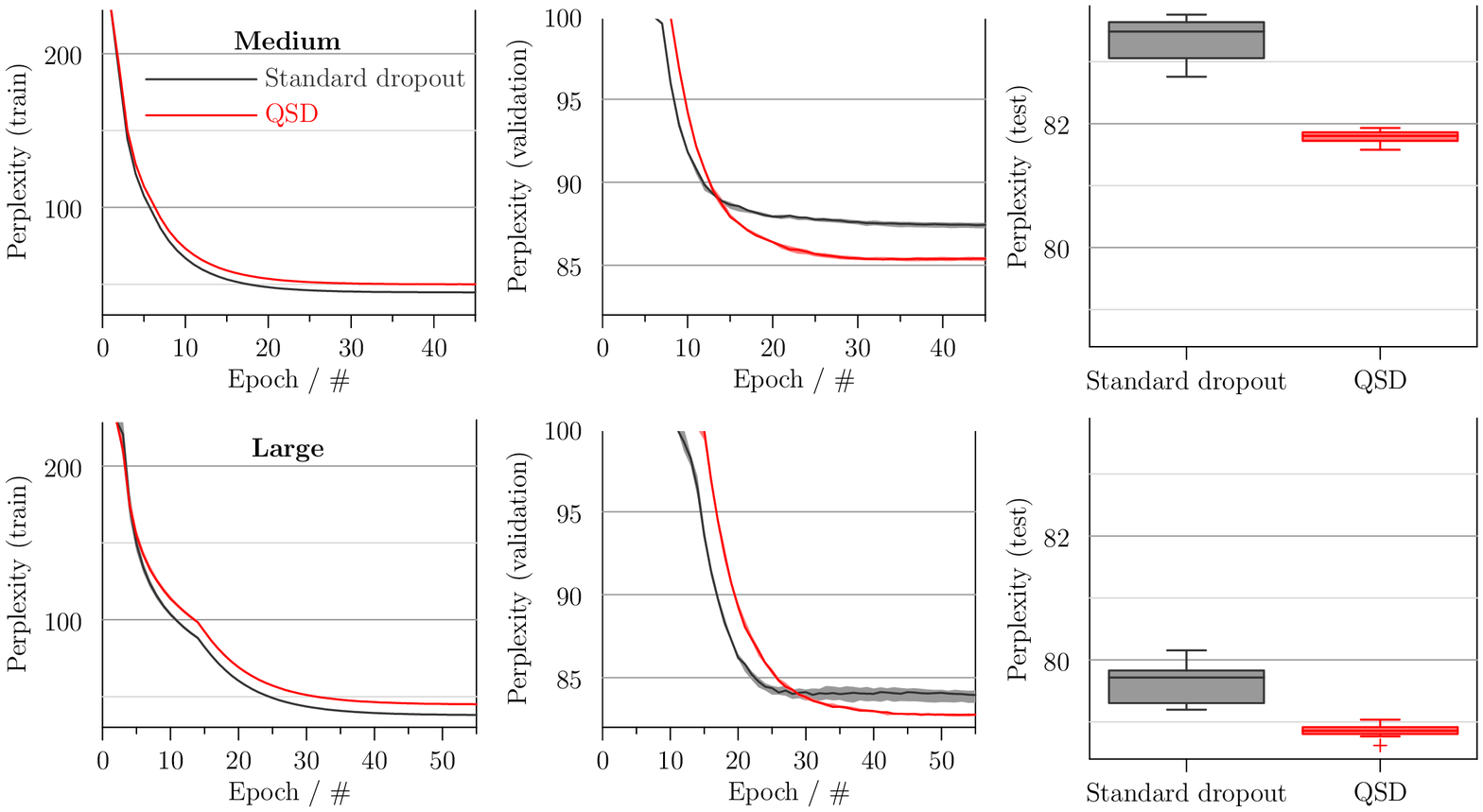} 

	\legend[fig:ptblstm]{Perplexity performances for the Penn Treebank language task with a medium- (\textit{top}) and
	large- (\textit{bottom}) sized LSTM model with standard dropout or QSD at non-recurrent connections (training
	perplexity inter-quartile range bands are shaded but are very small).}

\end{portrait}

The final comparison replicates the `naive dropout' condition from the set of sentiment analysis experiments of
\citet{gal2016theoretically} on the Cornell Film Review dataset of 5000 reviews \citep{pang2005seeing}.  Training
parameters were unchanged from the study of \citet{gal2016theoretically} with inputs of consecutive segments of 200
words, an embedding layer of 128, a 128-unit LSTM or GRU, followed by a fully connected single unit. A mean square error
cost function with an L2 regularisation coefficient of 0.001 was minimised using a ADAM optimiser with a batch size of
128 and learning rate of 0.001 for 100 epochs. As for \citet{gal2016theoretically}, the train/test split was set to
80\%/20\% and redistributed for every seed. Standard dropout or QSD was applied to non-recurrent connections with a
dropout rate of 0.5 and the QSD homogeneity $\alpha$ set to 1, resulting in a uniform distribution for quantal
heterogeneity, with no hyperparameter search. 

The descents in training error (Figure \ref{fig:cfrrnn}) are similar for standard dropout and QSD. For the LSTM model,
the final median error for QSD of 0.152 is less than that for standard dropout (0.161, $Z=2.94$, $P=0.003$). Similar
results were obtained for the GRU model (standard dropout 0.153, QSD 0.148, $Z=2.10$, $P=0.036$). Notably, the GRU QSD
test error at 100 epochs (0.148) approximates to the result of $\sim0.150$ obtained after training a variational LSTM
for 1000 epochs \citep{gal2016theoretically}.

\begin{portrait}[!htp] 
	
	\graphic[width=1.2\defwidth,trim={0 0 0 0},clip]{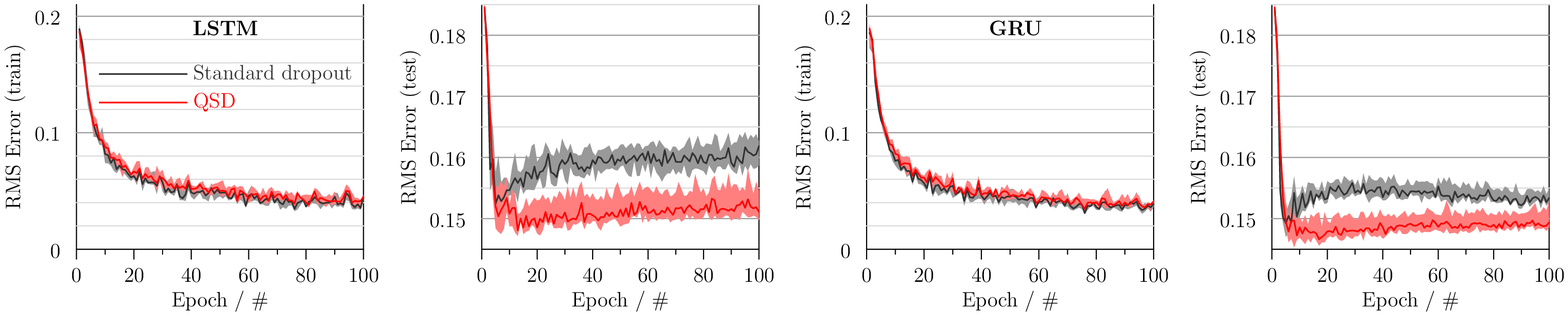} 

	\legend[fig:cfrrnn]{Error performances for the Cornell Film Review sentiment analysis task using an LSTM
	(\textit{left}) and GRU (\textit{right}) recurrent network with standard dropout or QSD at non-recurrent connections.}

\end{portrait}

\section{Conclusions}

QSD is a novel form of deep learning regularisation based on the quantal properties of biological synapses. It enhances
sparse encoding within hidden layers with reductions, compared with standard dropout, in the mean and variance of
outputs at test time without shifts in trainable weights and bias distributions. While we have reported these reductions
for the MLP experiments, similar findings were observed the other models. For example, mean final-ReLU layer activations
were less for QSD by 8\% in the LeNet-5 CNNs, and 2\% in the Wide ResNet CNNs (despite widespread batch-normalisation).
RNN outputs were reduced in the 1-layer LSTMs by 15\% for the mean absolute output (34\% by variance) and in the 2-layer
LSTMs by 17\% (38\%) by variance.

QSD introduces an additional hyperparameter $\alpha$ that is sensitive to the dropout rate. A consistent observation
throughout the experiments of the present study is that training with larger dropout rates is improved by increases in
the homogeneity parameter $\alpha$. Detailed theoretical explanation for this heuristic is desirable, but the optimal
choice is likely to be model- and/or data- dependent. Since QSD models synaptic behaviour, it could be argued that its
application to synaptic weights, similar to DropConnect, rather than outputs is more biological plausible. While
superficially attractive, this argument is dubious since it ignores the the role of beta distribution in modelling
homogeneous release probabilities: this probability distribution are not directly relevant to dendritic inputs but only
comes into effect in response to action potentials outputs propagated via axons to nerve terminals. Therefore fidelity
to biological phenomena is greater when applying to QSD to dropout rather than DropConnect.

While improved regularisation has been demonstrated for multilayer perceptrons, convolutional networks, and recurrent
networks there is scope for implementing QSD into more specialised applications. For example, convolutional networks
might benefit from applying QSD at a feature map level of granularity as employed for spatial dropout
\citep{tompson2015efficient}. Alternatively an activity-dependent component similar to Maxout
\citep{goodfellow2013maxout} could be introduced that incorporates kinetic models of short-term synaptic plasticity
\citep{tsodyks1997neural}. For recurrent networks, regularisation of recurrent connections \citep{gal2016theoretically}
using QSD may deliver superior performance.

\section{Appendix}

\subsection{Background on quantal synaptic transmission}

Quantal transmission at synapses is a result of the release of neurotransmitter molecules packaged inside discrete
membrane-bound vesicles from the pre-synaptic neuron. Upon release, the neurotransmitter activates receptors located on
the post-synaptic membrane resulting in a change in ionic conductance and evoking a synaptic current in the
post-synaptic neuron in response. The magnitude of the response in the post-synaptic neuron to a single vesicle
corresponds to the \textit{quantal size} $q$. The \textit{number of release sites} $n$ equates to the number of such
contacts between the pre-synaptic and post-synaptic neuron, and in the central nervous system $n$ may range from one to
many hundreds \citep{meyer2001estimation}. 

The instruction for the pre-synaptic neuron to release neurotransmitter is given by the arrival of nerve impulses that
are propagated by its axon from the cell body to the pre-synaptic terminals. Each nerve impulse, or action potential,
stimulates a voltage-activated influx of calcium into the pre-synaptic terminal. The influx induces a calcium-dependent
exocytotic fusion of a pre-synaptic vesicle with the membrane of the pre-synaptic neuron. This results in release of the
neurotransmitter contained inside vesicle into the synaptic cleft between the pre-synaptic and post-synaptic neurons.
However, the calcium-dependent exocytotic fusion is not deterministic since it is only occurs for a proportion of nerve
impulses. This proportion corresponds to the \textit{probability of release} $p$.

Consider a synapse in which the post-synaptic response $x$ to a single nerve impulse in the pre-synaptic neuron is
modelled with respect to the number of release sites $n$, quantal size $q$, and probability of release $p$. In the
simplest case, the probability of release $p$ at each site conforms to a Bernoulli distribution that is identical for
all $n$ sites.  The distribution of the number of Bernoulli successes $i$ is described by the binomial probability mass
function $\mathcal{B}(i \vert n, p)$:

\begin{align}
 \mathcal{B}(i \vert n, p) &= \frac{n!}{i!(n-i)!} p^i(1-p)^{n-i} , &0 \le i \le n, 0 \le p \le 1
\end{align}

The distribution of responses $x$ is therefore determined by a binomial probability mass function
\citep{kuno1971quantum} according to the linear relation $x \propto i$; the constant of proportionality defines the
quantal size $q$: $x = iq$. At a given probability of success $p$, the first moment of a binomial probability mass
function $\mathcal{B}(i \vert n, p$) is $np$. We can thus express the expectation for the mean response $\mathbb{E}[x]$
as a product that scales the first moment of the binomial probability mass function by the quantal size $q$.

\begin{align}
	\mathbb{E}[x] &= npq \label{eq:mnpq}
\end{align}

In this simple case, the binomial model assumes a constant quantal size $q$ and number of release sites $n$ independent
of a fixed probability of release $p$. Experimental recordings from central neurons show however that synapses exhibit
heterogeneities in the probabilities of release $p$ \citep{walmsley1988nonuniform,murthy1997heterogeneous} and quantal
sizes $q$ \citep{lisman1993quantal,bekkers1994quantal}. The binomial model therefore represents a simplification of
quantal synaptic transmission at central synapses.

For more detailed accounts of mathematical models of quantal synaptic transmission, the reader is referred to somewhat
more recent work on quantal analysis \citep{silver2003estimation,bhumbra2013reliable}. Quantal synaptic dilution is
based on modelling distributed values for the two quantal parameters: probability of release $p$ and quantal size $q$.
The technique combines both heterogeneities modelling dropout retain probabilities as analogous to release
probabilities, and rescaling sizes representing quantal sizes.

\subsection{Quantal synaptic dilution}

`Diluted connectivity' is a term used in neuroscience to quantify the partial connectivity between two brain areas. For
example, within the CA3 region of the hippocampus, the degree of connectivity between neurons can be less that 10\%
\citep{rolls2012advantages}. This is an example of `strong dilution' in which the partial connectivity is permanent,
whereas in `weak dilution' a fixed proportion of connections are transiently removed at random
\citep{hertz1991introduction}. Like standard dropout \citep{hinton2012improving,srivastava2014dropout}, quantal synaptic
dilution is a form of weak dilution.

Quantal synaptic dilution (QSD) is applied to the activations $\mathbf{a}$ of a layer of $n$ units with a transfer
function $g$, trainable weights $\mathbf{W}$ and biases $\mathbf{b}$, receiving input $\mathbf{x}$ given by:

\begin{align}
	\mathbf{a} &= g(\mathbf{Wx}+ \mathbf{b}) \label{eq:act}
\end{align}

Also like standard dropout, QSD includes a dropout hyperparameter $d$ that determines the mean retain probability
$\bar{p} = 1 - d$. In order to model heterogeneities in retain probabilities, QSD includes an additional hyperparameter
$\alpha$, where an increase in value results in greater homogeneity tending towards standard dropout as $\alpha \to
\infty$.  Heterogeneity in retain probabilities $p$ is modelled using a beta probability density function $\mathcal{P}(p
\vert \alpha, \beta)$ where $\beta$ is determined by the functional relation: $\beta = \alpha (1-\bar{p})/\bar{p}$. 

The choice of a beta distribution is based on the use of a beta probability density function used to model heterogeneity
of release probabilities at central synapses \citep{silver2003estimation,bhumbra2013reliable}. While the distribution of
release probabilities at hippocampal synapses were fitted initially with the closely related gamma probability density
function \citep{murthy1997heterogeneous}, the advantage of using a beta distribution for probabilities $p$ is that its
support set is defined over the range $[0,1]$: 

\begin{align}
\mathcal{P}(p \vert \alpha, \beta) &= \frac{p^{\alpha-1}(1-p)^{\beta-1}}{B(\alpha, \beta)},
	\phantom{i}\alpha, \beta > 0,
	\phantom{i}B(\alpha, \beta) = \frac{\Gamma(\alpha)\Gamma(\beta)}{\Gamma(\alpha+\beta)}, 
	\phantom{i}\Gamma(\gamma) = \int_{0}^{\infty} g^{\gamma-1}e^{-g}dg
\end{align}

If $\mathbf{m}$ denotes a Boolean mask vector, where $m_i$ is the $i$th Bernoulli sample with the retain probability
$p_i$, its sample space represents discrete realisation of heterogeneity in retain probabilities: 

\begin{align}
	m_i \sim \textrm{Bernoulli}(m \vert p_i),\phantom{M} &p_i \sim \mathcal{P}(p \vert \alpha, \beta)
\end{align}

The expected mean of the Boolean masks $m_i$ corresponds to the mean retain probability $\bar{p}$, and therefore direct
Hadamard multiplication with the activations $\mathbf{a}$ without rescaling would result in a reduction in mean
activations at training time. If $\mathbf{q}$ denotes a vector of rescaling sizes, each element $q_i$ is multipled by
the corresponding boolean mask value $m_i$ to express a vector $\mathbf{c}$ of coefficients $\{c_0, c_1, \ldots, c_n\}$
to multiply element-wise with the activations $\mathbf{a}$ during training.

\begin{align}
	\mathbf{y} &= \mathbf{c} \odot \mathbf{a},
	\phantom{i}c_i = q_i m_i,
	\phantom{i}m_i \sim \textrm{Bernoulli}(m \vert p_i),
	\phantom{i}p_i \sim \mathcal{P} (p \vert \alpha, \beta )
	\label{eq:qsd_0}
\end{align}

In QSD, heterogeniety in the rescaling sizes $\mathbf{q}$ is based on its linear relation with the beta distributed retain
probabilities $\mathbf{p}$. The linear relation between the retain probabilities $\mathbf{p}$ and rescaling sizes
$\mathbf{q}$ is consistent with biological observations since strong positive correlations between release probabilities
and quantal sizes have been demonstrated at synapses of pyramidal cells in the visual cortex
\citep{hardingham2010quantal} with correlation coefficients as high as $r=0.89$ in layer V. 

As for the commonly used `inverted' implementation of standard dropout, scalar division is performed to establish
correspondence between the expected mean for the rescaling sizes and the reciprocal of the mean retain probability
$1/\bar{p}$. The rescaling element $q_i$ for the $i$th unit is evaluated by dividing its corresponding retain
probability $p_i$ by the square of the mean retain probability $q_i = p_i/(\bar{p})^2$, giving an expected mean for
$\mathbf{q}$ of $1/\bar{p}$.  Combining the heterogeneity of rescaling sizes with Equation \ref{eq:qsd_0}, and
expressing $\beta$ with respect to its functional relationship with $\alpha$ and $\bar{p}$, gives:

\begin{align}
	\mathbf{y} &= \mathbf{c} \odot \mathbf{a},
	\phantom{i}c_i = q_i m_i,
	\phantom{i}q_i = \frac{p_i}{(\bar{p})^2},
	\phantom{i}m_i \sim \textrm{Bernoulli}(m \vert p_i),
	\phantom{i}p_i \sim \mathcal{P}\left(p \vert \alpha, \alpha \left( \frac{1-\bar{p}}{\bar{p}} \right)\right)
\end{align}

While QSD was implemented for the present study within the TensorFlow framework \citep{tensorflow2015-whitepaper}, the
library presently has no random number generator for beta-distributed sampling. Since a random number generator for
gamma-distributed samples is included, two independent gamma distributed random variables $\bm{\gamma}$ and
$\bm{\lambda}$ can be sampled with gamma probability density functions:

\begin{align}
	f(\bm{\gamma } \vert \alpha) &= \frac{1}{\Gamma(\alpha)}  \gamma^{\alpha-1}e^{-\gamma }, \phantom{M}\gamma,  \alpha > 0\\
	f(\bm{\lambda} \vert \beta ) &= \frac{1}{\Gamma(\beta )} \lambda^{\beta -1}e^{-\lambda}, \phantom{M}\lambda, \beta > 0
\end{align}

If the two gamma variables are sampled independently, the following relation \citep{papoulis2002probability} can be used
to sample a beta-distributed variable $p$ with parameters $\alpha$ and $\beta$.

\begin{align}
	p_i &= \frac{\gamma_i}{\gamma_i + \lambda_i}, \phantom{M}p_i \sim \mathcal{P}(p \vert \alpha, \beta) =  \frac{p^{\alpha-1}(1-p)^{\beta-1}}{B(\alpha, \beta)},
\end{align}

A Python implementation of QSD for TensorFlow therefore occupies only a few lines of code:

\begin{lstlisting}
import tensorflow as tf
from tensorflow.python.ops import math_ops, random_ops

def qsd(activations, bernoulli_shape, drop_rate, alpha):
  keep_prob = 1 - drop_rate
  beta = alpha * drop_rate / keep_prob
  gamma = tf.random.gamma(bernoulli_shape, alpha, 1)
  lmbda = tf.random.gamma(bernoulli_shape, beta, 1)
  retain_prob = gamma / (gamma + lmbda)
  rescale_size = retain_prob / (keep_prob * keep_prob)
  random_uniform = random_ops.random_uniform(bernoulli_shape,\
	                                     dtype=x.dtype)
  keep_mask = math_ops.cast(retain_prob > random_uniform,\
	                    activations.dtype)
  coefficients = keep_mask * rescale_size
  return activations * coefficients
\end{lstlisting}

\subsection{Rescaling effects during training}

Given an expected mean for the retain probabilities $\mathbf{p}$ of $\bar{p}$ and for rescaling sizes $\mathbf{q}$ of $1
/ \bar{p}$, one might assume the expected mean for their product used to evaluate the coefficients $\mathbf{c}$
approximates to one. The linear relation between $\mathbf{p}$ and $\mathbf{q}$ however introduces covariances between
the two variables that would increase the average of the coefficients above one. The extent of the increase depends on
the choice of the dropout rate $d$ and homogeneity $\alpha$. In this section, we will show that the lower bound for the
average of the coefficients is one and its value increases with reductions in the homogeneity parameter $\alpha$ and
increases in dropout rate. 

The coefficients in $\mathbf{c}$ are effectively drawn from the same underlying beta distribution that is sampled twice
and multiplied by one another. In the first case, the beta distribution is used to evaluate the retain probabilities for
sampling the boolean Bernoulli masks $\mathbf{m}$. The mask values are then multiplied by the rescaling sizes
$\mathbf{q}$, that are evaluated by dividing the same beta distribution by a scalar corresponding to the square of its
expectation $\mathbb{E}[p]$. Beyond the division, the continuous functional form of the beta distribution is therefore
effectively multiplied by its own discretely realised counterpart. Since discretisation does not affect the expectation
of the mask distribution, the expectation of the coefficients $\mathbb{E}[c]$ equates to the expectation of the squared
retain probabilities, scaled by the inverse of the squared expectation of retain probabilities:

\begin{align}
	\mathbb{E}[c] &= \frac{1}{\mathbb{E}[p]^2} \mathbb{E}[p^2]\\
	&= \left( \frac{\alpha + \beta}{\alpha} \right)^2 \int_0^1 p^2 \frac{p^{\alpha-1}(1-p)^{\beta-1}}{B(\alpha, \beta)} dp
\end{align}

The beta function can be expressed with respect to the quotient of gamma functions:

\begin{align}
	\mathbb{E}[c]
	&= \left( \frac{\alpha + \beta}{\alpha} \right)^2 \int_0^1 p^2 
	   \frac{\Gamma(\alpha + \beta)}{\Gamma(\alpha) \Gamma(\beta)} p^{\alpha-1}(1-p)^{\beta-1} dp\\
	&= \left( \frac{\alpha + \beta}{\alpha} \right)^2 \frac{\Gamma(\alpha + \beta)}{\Gamma(\alpha) \Gamma(\beta)}
	   \int_0^1 p^{\alpha+1}(1-p)^{\beta-1} dp\\
	&= \left( \frac{\alpha + \beta}{\alpha} \right)^2 
		 \left( \frac{\Gamma(\alpha + \beta)}{\Gamma(\alpha) \Gamma(\beta)} \right)
		 \left( \frac{\Gamma(\alpha+2)\Gamma(\beta)}{\Gamma((\alpha +2) + \beta)} \right)
		 \int_0^1 \frac{\Gamma((\alpha +2) + \beta)}{\Gamma(\alpha+2)\Gamma(\beta)} p^{(\alpha+2)-1}(1-p)^{\beta-1} dp
\end{align}

The last step above is useful because the integration now corresponds to a definite integral, over a properly normalised
beta probability density function $\mathcal{P}(p \vert (\alpha+2), \beta)$, that integrates to one and can thus be
eliminated.

\begin{align}
	\mathbb{E}[c]
	&= \left( \frac{\alpha + \beta}{\alpha} \right)^2 
	\frac{\Gamma(\alpha+2)\Gamma(\alpha + \beta)}{\Gamma(\alpha+\beta+2)\Gamma(\alpha)}
\end{align}

With repeated use of the recurrence relation $\Gamma (\gamma + 1) = \gamma \Gamma (\gamma)$, the right hand expression
can be further simplified:

\begin{align}
	\mathbb{E}[c]
	&= \left( \frac{\alpha + \beta}{\alpha} \right)^2
	\frac{(\alpha+1)\Gamma(\alpha+1)\Gamma(\alpha + \beta)}{(\alpha+\beta+1)\Gamma(\alpha+\beta+1)\Gamma(\alpha)}\\
	&= \left( \frac{\alpha + \beta}{\alpha} \right)^2
	\frac{\alpha (\alpha+1)\Gamma(\alpha)\Gamma(\alpha + \beta)}{(\alpha + \beta)(\alpha+\beta+1)\Gamma(\alpha+\beta)\Gamma(\alpha)}\\
	&= \left( \frac{\alpha + \beta}{\alpha} \right) \left( \frac{\alpha+1}{\alpha+\beta+1} \right)\\
	&= \frac{ \alpha (\alpha + \beta + 1) + \beta}{\alpha(\alpha + \beta + 1)}\\
	\therefore \mathbb{E}[c]	&= 1 + \frac{\beta}{\alpha (\alpha + \beta + 1)} \label{eq:coef}
\end{align}

Since $\alpha,\beta>0$, Equation \ref{eq:coef} proves that lower bound for the expectation of the coefficients is 1.
Equation \ref{eq:coef} also shows that the expectation of the coefficients would increase with reductions in $\alpha$
and increases in $\beta$ (and therefore dropout rate). These relations are illustrated in Figure \ref{fig:exp_coef}
(this document) for different values of $\alpha$. 

\begin{portrait}[h] 
	
\graphic[width=0.5\defwidth,trim={0cm 0cm 0cm 0.5cm}]{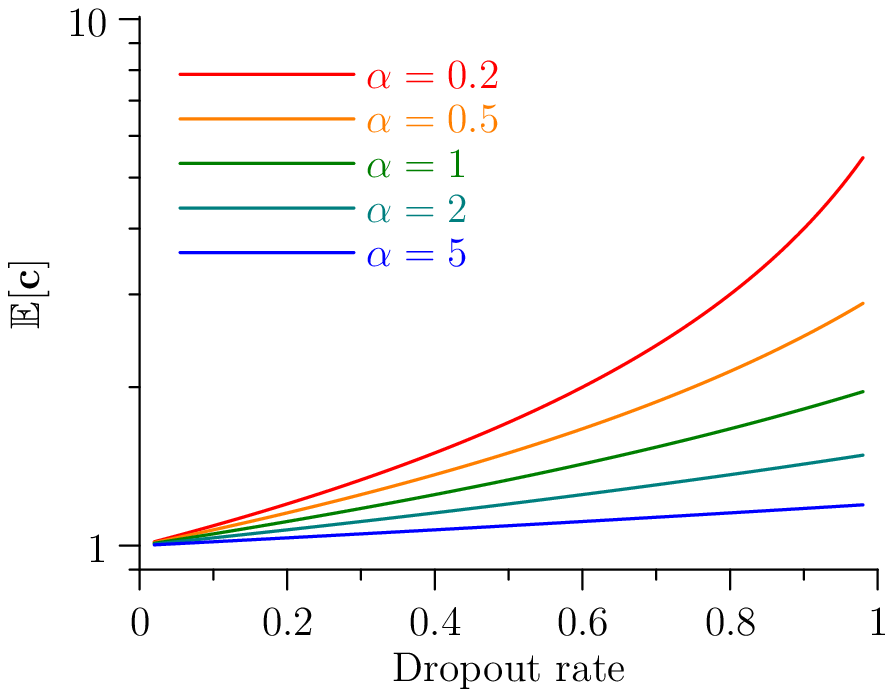} 

	\legend[fig:exp_coef]{Coefficient expectations plotted against dropout rate for different values for $\alpha$.} 

\end{portrait}

While we have presented a theoretical treatment of modelling the expectation of the coefficients $\mathbf{c}$, the
results have been confirmed empirically by numerical computation. It is possible that the regularisation performance of
QSD could be further improved by adjusting the scaling of the rescaling sizes $q$ to ensure an a coefficient expectation
of one, $\mathbb{E}[c]=1$, in accordance to this analytical solution. This possibility is under investigation for the
future.

\subsection{Effects on activations after training}

The main manuscript describes a number of experiments that compare QSD with standard dropout. This section gives more
detailed consideration of the effects of QSD on hidden layer outputs for the experiments on a fully-connected multilayer
perceptron network trained on the MNIST data set (60000 training and 10000 test samples of 28x28 images with 10
classes). The network is composed of three hidden layers of 128, 256, and 512 ReLU units with a softmax 10-unit output
layer.

Cross-entropy costs were minimised using a stochastic gradient descent optimiser with momentum $(\mu=0.9)$ and a batch
size of 64 for 100 epochs. The initial learning rate of 0.01 was multiplied by 0.2 on epochs 30, 60, then 80.  Weights
were initialised using variance scaling \citep{he2015delving}.  Dropout or QSD masks were positioned after each of the
three ReLU non-linearities at identical rates. No other regularisation or data augmentation was used.

In addition, in order to isolate the contributions of distributed retain probabilities and distributed rescaling sizes,
we fixed either the rescaling size $q$ to $1/\bar{p}$ or the retain probability $p$ to $\bar{p}$, while maintaining the
distributed heterogeneity in the other parameter. For $\alpha$ values of 0.2 and 0.5, the learning curves (Figure
\ref{fig:mnistmlppqsup}A, this document) demonstrate superior test performance of QSD over standard dropout and over
both the two isolated distributed conditions. 

\begin{portrait}[!htp] 
	
	\graphic[width=1.5\defwidth,trim={0 0 0 0},clip]{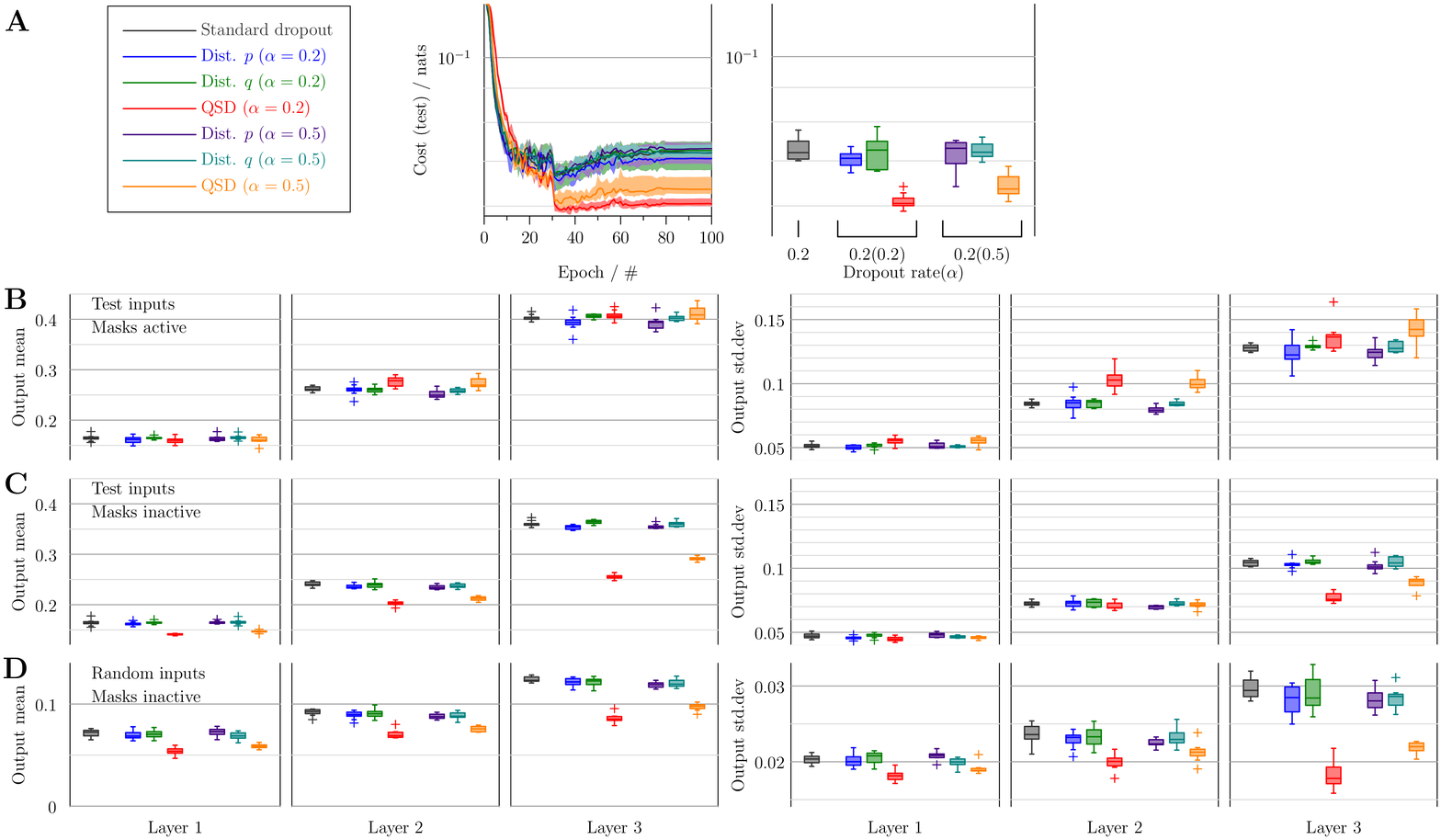} 

	\legend[fig:mnistmlppqsup]{Comparisons of standard dropout, distributed retain probabilities (Dist. $p$), distributed
	quantal sizes (Dist. $q$), and QSD. \textbf{A:} Learning curves and test performances for the different conditions.
	\textbf{B}: Box and whiskers plots summarising the outputs following training from 8 runs for each of the different
	conditions averaging over all units within each of the 3 hidden layers showing the mean (\textit{left}) and standard
	deviation (\textit{right}) across all samples, forward passing test inputs with the dropout masks active \textbf{C}:
	The same outputs for a forward pass of the test data at test time with masks inactive.  \textbf{D}: The same outputs
	forward passing the randomised data with masks inactive (note reduced $y$-axes scales). }

\end{portrait}

The histograms shown in Figure 3 of the main manuscript gives examples of output distributions for each of the layers
for the different conditions but does not summarise the data across all runs. Figures \ref{fig:mnistmlppqsup}B-D (this
document) shows box and whiskers plots summarising the group data from all 8 runs for the outputs averaged over all
units following training for each of the different conditions. The group data was obtained for each of the 3 hidden
layers and Figures \ref{fig:mnistmlppqsup}B-D illustrate the group distributions for the means (left) and standard
deviations (right) across all 10000 test samples. For Figure \ref{fig:mnistmlppqsup}B, the dropout masks were kept
active as during training.  Comparisons between the different conditions show no systematic differences across all
hidden layers.

For a good sparse model there should only be a few number of highly activated units \citep{glorot2011deep}, with the
output of hidden layers exhibiting a low mean and variance of activations across different data samples
\citep{srivastava2014dropout}. When evaluating the outputs shown in Figure \ref{fig:mnistmlppqsup}C, masks were inactivated
as during test time. The output across units, particularly in the third layer, show a reduction in mean activation for
the QSD layers compared to the other conditions as well as a reduction in standard deviation. The relative progression
of the reduction in mean activation across layers suggests that QSD facilitates propagation of sparse encoding along the
network.

Figure \ref{fig:mnistmlppqsup}D illustrates similar box and whisker plots with the masks inactive but with the input
pixels of the test data randomly permutated to remove any meaningful signal while preserving the original input
distribution. Only the QSD layers systematically exhibit increased sparsity across units with relative reductions in
mean and standard deviation of outputs across samples compared to standard dropout. Once again the reduction in outputs
for QSD along the network compared to the other conditions suggests a facilitated propagation of sparse encoding, with a
minimal activation of the final hidden layer that is most compatible with an encoding of random inputs. Improvement by
QSD in sparse encoding and network regularisation must therefore result from a dilution during training that combines
heterogeneous retain probabilities with distributed rescaling sizes.

\bibliography{qsd}

\end{document}